\documentclass{article} 
\usepackage{colm2024_conference}

\usepackage{microtype}
\usepackage{hyperref}
\usepackage{url}
\usepackage{booktabs}
\usepackage{graphicx}
\usepackage{setspace}
\usepackage{amsmath}
\usepackage{amsthm}
\usepackage{caption}
\usepackage{subcaption}
\usepackage{booktabs}

\theoremstyle{definition}
\newtheorem{definition}{Definition}[section]

\title{Dictionary Learning Improves Patch-Free Circuit Discovery in Mechanistic Interpretability: A Case Study on Othello-GPT}

\colmfinalcopy

\author{Zhengfu He$^{1}$\hspace{.3em}
Xuyang Ge$^{1}$\hspace{.3em}
Qiong Tang$^{1}$\hspace{.3em}
Tianxiang Sun$^{1}$\hspace{.3em}
Qinyuan Cheng$^{1}$\hspace{.3em}
Xipeng Qiu$^{1}$
\\
\\
\texttt{zfhe19@fudan.edu.cn} \\
\\
$^{1}$OpenMOSS Team, Fudan University
}

\fancypagestyle{firstpage}{
  \lhead{\begin{picture}(0,0)\put(0,-6){\includegraphics[width=0.3\linewidth]{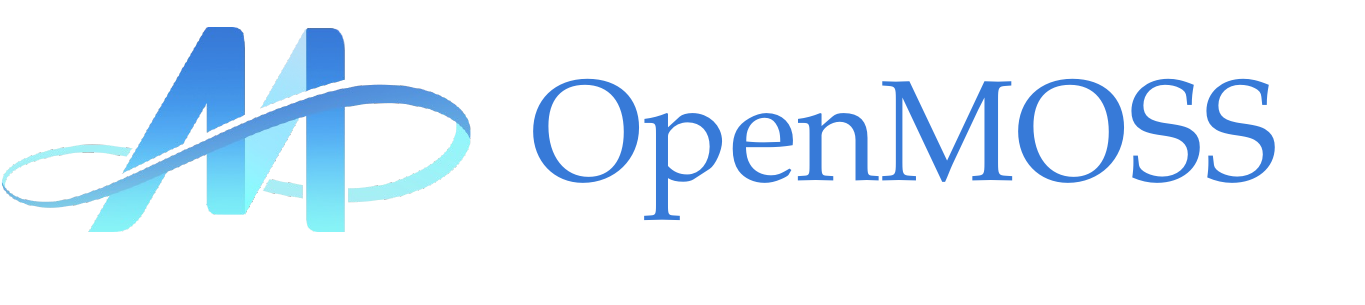}}\end{picture}}
}

\begin{document}

\maketitle

\thispagestyle{firstpage}

\begin{abstract}
Sparse dictionary learning has been a rapidly growing technique in mechanistic interpretability to attack superposition and extract more human-understandable features from model activations. We ask a further question based on the extracted more monosemantic features: \textbf{How do we recognize circuits connecting the enormous amount of dictionary features?} We propose a circuit discovery framework alternative to activation patching. Our framework suffers less from out-of-distribution and proves to be more efficient in terms of asymptotic complexity. The basic unit in our framework is dictionary features decomposed from all modules writing to the residual stream, including embedding, attention output and MLP output. Starting from any logit, dictionary feature or attention score, we manage to trace down to lower-level dictionary features of all tokens and compute their contribution to these more interpretable and local model behaviors. We dig in a small transformer trained on a synthetic task named Othello and find a number of human-understandable fine-grained circuits inside of it.
\end{abstract}




\section{Introduction}

In recent years, advances in transformer-based language models\citep{Vaswani2017Transformer,Brown2020GPT3} have sparked interest in better understanding the internal computational workings of these systems. 
Researchers have made some progress in identifying interpretable circuits and algorithms within GPT-2 level models\citep{Olsson2022induction,Wang2023IOI}, but much of the models' broad language generation capabilities remain opaque. 
The emerging field of \textit{Mechanistic Interpretability}\citep{Cammarata2020distillcircuitthread,Elhage2021mathematical} aims to reverse engineer neural networks in order to map their internal components to understandable computational primitives. 
By decomposing these black box systems into basic building blocks carrying out human-understandable functions, the goal is to shed light on how complex behaviors like language modeling emerge from combinations of simple computational elements.

One main obstacle from fully understanding neural networks is the superposition hypothesis\citep{Arora2018LinearStructure,Elhage2022superposition}.
The superposition hypothesis assumes that the model learns to represent more linear features than it has hidden dimensions.
In the research agenda of mechanistic interpretability, understanding model activations by extracting human-understandable components out of superposition is a central task.
Recent advances in sparse dictionary learning\citep{Chen2017dict,Subramanian2018dict,Zhang2019dict,Panigrahi2019dict,Yun2021dict,Cunningham2023dict} have opened up new possibilities for extracting more interpretable, monosemantic features out of superposition. 
By learning sparse dictionaries that decompose activations into semantically meaningful directions in the representation space, we are able to gain more microscopic insight of model representations.

Circuit analysis\citep{Conmy2023ACDC,Olah2020zoom} aims to find functional connections between internal features. The current approach to extracting circuits relies much on \textit{Activation Patching} or ablation \citep{Wang2023IOI,Conmy2023ACDC,Zhang2023Bestpractice}. They have shown impressive results in understanding both attention and MLP module, but still they are faced with some problems e.g. out-of-distribution\citep{Zhang2023Bestpractice} and the hydra effect\citep{Wang2023IOI,McGrath2023Hydra}. Patching-based methods also suffers from quadratic asymptotic complexity. This work presents a novel alternative to extracting interpretable circuits connecting dictionary features without patching.

We trained dictionaries on the output of both attention and MLP layers of a model solving a synthetic task named Othello. Thanks to the massive linear structure of Transformers\citep{Elhage2021mathematical}, all dictionary features (attention scores) can be decomposed into contributions of lower-level dictionary features (feature pairs). In this way, we can trace all the way down from any logit or any given dictionary feature in an iterative way to find meaningful circuits causing it activated.

Our main contributions are as follows.
\begin{itemize}
    \item We outline several choices of positions to train dictionary on and make a comprehensive comparison among them. We claim it preferable to decompose the output of each module writing to the residual stream e.g. word embedding, attention output and MLP output.
    \item (Minor) We use dictionary learning to find interpretable features inside of an Othello model. The result shows that dictionary learning automatically finds \textit{all} types of features discovered by probing, in an unsupervised manner.
    \item We exploit our circuit discovery theory to find meaningful subgraph of the whole computational graph of the Othello model. The experimental results reveal a large partition of the inner information flow of the model, turning the community's microscopic understanding of the Othello model to macroscopic one. To our best knowledge, this is the first work to locate interpretable circuits in a patch-free manner.
\end{itemize}

\section{Patch-Free Circuit Discovery with Dictionary Learning}

\subsection{Interpretable Sparse Coding with Dictionary Learning}

\begin{figure}[h]
    \centering
    \includegraphics[width=\linewidth]{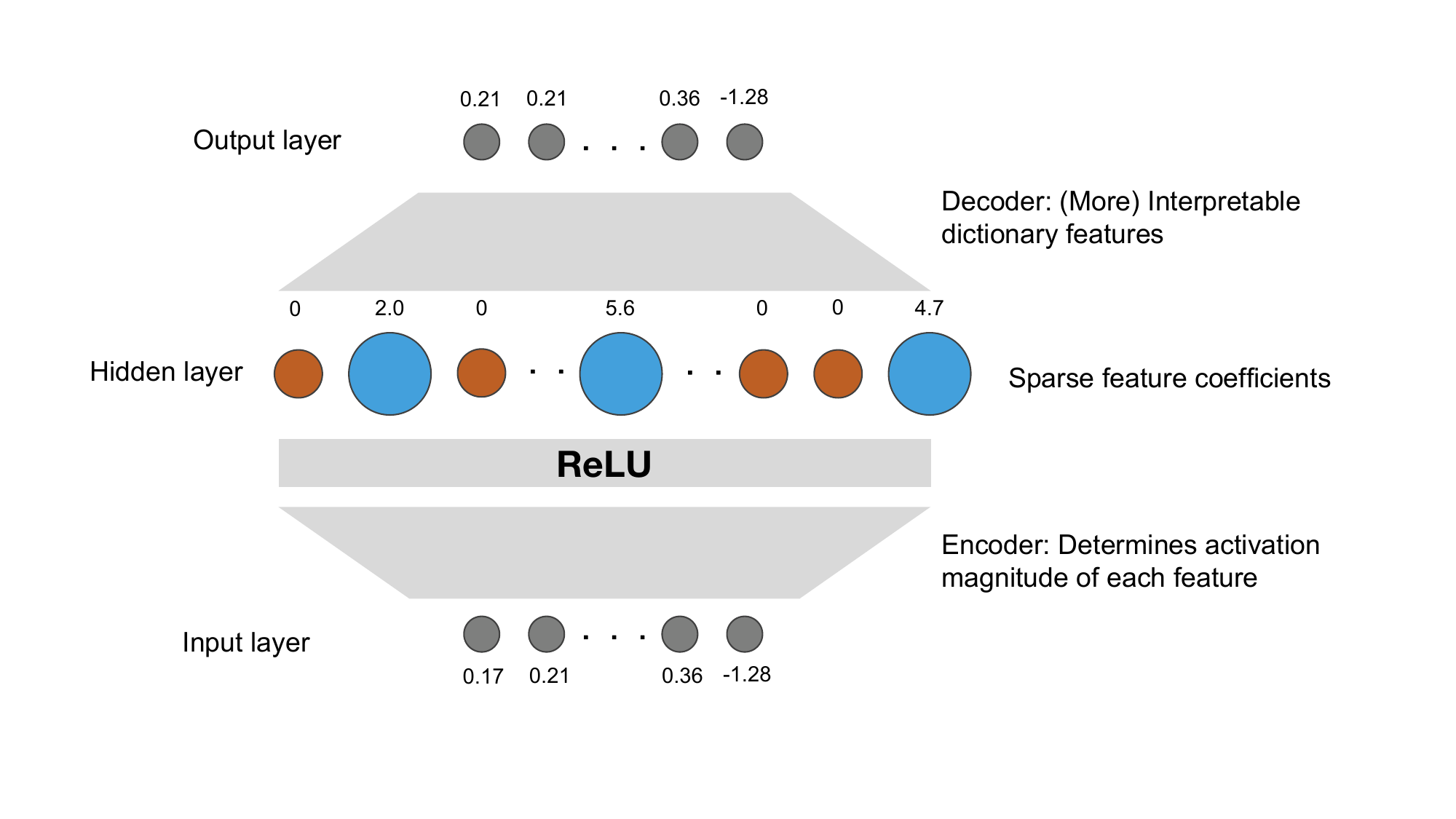}
    \caption{Activation of a given position and a given input decomposed into a sparse reconstruction of dictionary features.}
    \label{fig:Dictionary}
\end{figure}

The goal of sparse dictionary learning is to find, through an autoencoder, a set of overcomplete bases $\mathbf{d}$ such that for any activation $\mathbf{x}$ at a given activation space, it can be decomposed into a sparse weighted sum over this set of bases:

\begin{equation}
\label{equation: Dictionary_decomposition}
\mathbf{x} \approx \sum_{i}{w_i \mathbf{d}_i};\quad \text{s.t.} \quad \min{\lVert \{w_i | w_i\in \mathbf{w}, w_i > 0\} \rVert}.
\end{equation}

$w_i$ is the activation magnitude of the $i$-th dictionry feature and $\mathbf{d}_i$ is its corresponding unit vector representing the feature direction.
By constraining the sparsity of the activations over the dictionary features, the dictionary is forced to find the fundamental features implicitly contained in the representation and compute the most sparse (under a given metric) composition over this set of features that can reconstruct the given representation.

For a model activation captured at a given position in the transformer, we can decompose it into a weighted sum of a group of more monosemantic dictionary features, as shown in Figure~\ref{fig:Dictionary}.

\subsection{Where to Train Dictionaries on?}
\label{section: where to train?}
\begin{figure}[h!]
    \centering
    \includegraphics[width=\linewidth]{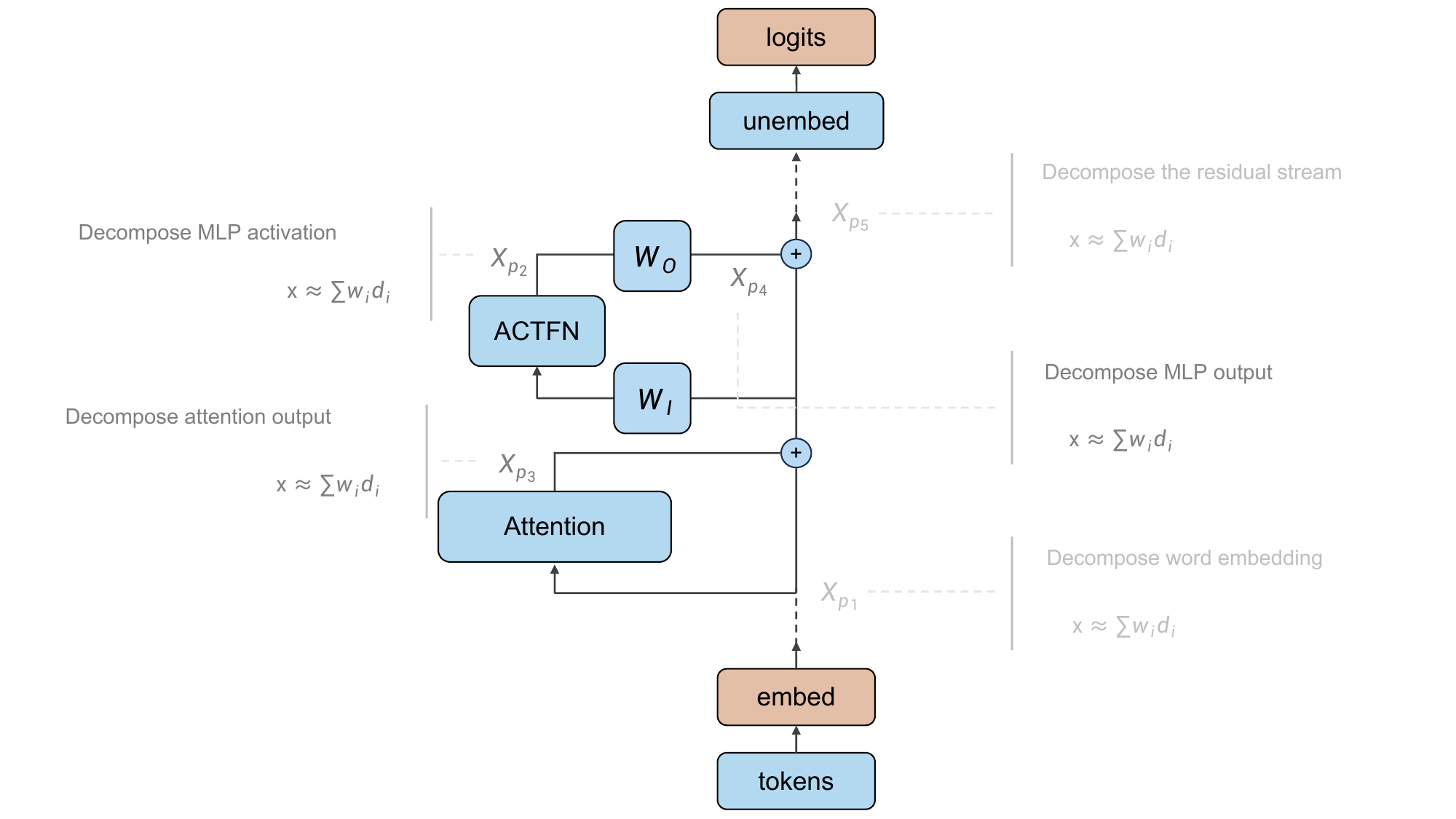}
    \caption{Five choices of positions to decompose via dictionary learning. Prior work has studied decomposing $X_{P1}$, $X_{p2}$ and $X_{p5}$, namely word embedding, MLP hidden layer and the residual stream. We claim it preferable to decompose $X_{P1}$, $X_{P3}$ and $X_{P4}$ i.e. word representations, the output of each attention layer, and the output of each MLP layer.}
    \label{fig:Decompose_position}
\end{figure}

As shown in Figure~\ref{fig:Decompose_position}, previous work based on dictionary learning has typically studied: word embedding\citep{Chen2017dict,Subramanian2018dict,Zhang2019dict,Panigrahi2019dict}, residual streams\citep{Yun2021dict,Cunningham2023dict}, and MLP hidden layers\citep{Bricken2023monosemanticity}. Here is a brief commentary on these works:
\begin{itemize}
    \item \textbf{Decomposing word representations}: Word representations are the initial state of the residual stream and trigger of subsequent computation. 
    Existing work that uses dictionary learning to decompose word representations into interpretable additive features is a good starting point. 
    However, this alone is not enough to fully understand the entire model.

    \item \textbf{Decomposing the residual stream}: Decomposing the residual stream between each transformer block has the benefit of being intuitive i.e. features computed by the first N layers. 
    However, this approach has some problems. If a shared dictionary is used to decompose all positions of the residual stream, distributional shifts at different model depth will greatly impact dictionary features.
    If each position is trained separately, the input and output will lie in two different vector spaces spanned by different dictionary bases, so their difference cannot be directly computed. 
    Corresponding concepts between the two bases would need to be identified to enable interpretation of circuits.

    \item \textbf{Decomposing MLP hidden layers}: This choice is based on the fact that in the prevalent pre-norm Transformer architecture, we only need to understand the components written to the residual stream by each module and sum them to obtain the final result, without needing to understanding the residual stream itself. 
    Under this view, we have two strategies: decomposing the MLP output or MLP hidden layers. 
    Compared to the former, directly analyzing the MLP hidden layers increases the number of trainable model parameters by about 4-16 times \footnote{(d\_MLP / d\_model) squared}.
    In addition, since the MLP down-projection matrix MLP\_out represents a linear mapping from a high-dim to low-dim space, some decomposed features could lie in its nullspace\citep{Nanda2023replication}.
\end{itemize}

Based on the above analysis, we believe it could be beneficial to use dictionary learning to decompose the following three parts: word representations, the output of each attention layer, and the output of each MLP layer.
Although there has already been considerable Mechanistic Interpretability work analyzing Attention heads compared to MLPs, we think incorporating them into a unified dictionary learning framework is necessary. A concurrent interim research report\citep{Kissane2024AttentionSAE} shares the same belief that this setting would be helpful for understanding Transformers in a systematic and scalable way.

\subsection{Identifying Information Flow in OV Circuits}
\label{section: theory_OV_circuit}

\begin{figure}[h!]
    \centering
    \includegraphics[width=\linewidth]{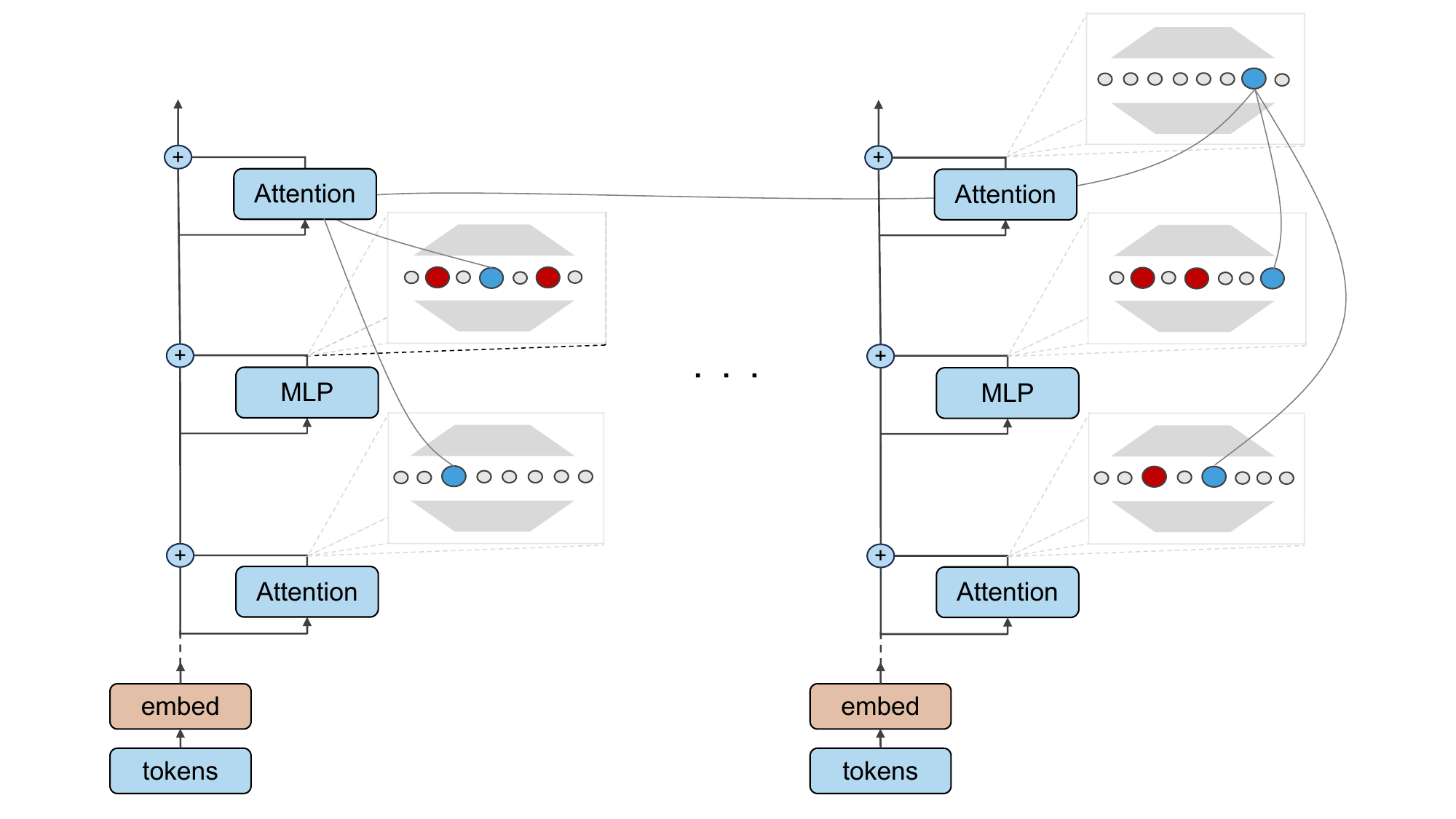}
    \caption{Attention features decomposed into contributions of lower-level dictionary features of previous tokens via the OV circuit.}
    \label{fig:OV_decomposition}
\end{figure}

In a Transformer model, each attention head needs to transfer the input of token $j$ to token $i$ $(j < i)$ through the following process:

\begin{equation}
    \textbf{OV}^h_{i \gets j} = \textbf{AttnPattern}^h_{ij}W_O^hW^h_Vx_j
\end{equation}

The superscript $h$ indicates that the corresponding model parameters/activations are independent for each head. 
$\textbf{AttnPattern}^h_{ij}$ represents the weight coefficients for token-wise mixing computed by the QK circuit. We treat it as constant here since its computational independence from the OV circuit. $W^h_O$ and $W^h_V$ are the OV weight matrices for this attention head.
$x_j$ is the input to this attention module at the residual stream of the $j$-th token, which is the same for each head so we omitted the superscript $h$.\footnote{We only consider the mainstream pre-norm architecture. Consequently, $x_j$ is the activation after the LayerNorm operation. We omit LayerNorm throughout the context and override the denotation of $x$ for both input and output of LayerNorm for simplicity. The rationale for this is that LayerNorm can be approximated as a linear operation. See Appendix~\ref{Appendix: sec: LayerNorm} for more discussion.}

Due to the independent additivity of multi-head attention, the output of the attention module of the $X$-th layer at token $i$ $\textbf{Out}_{\text{LXA},i}$can be expressed as shown in Equation~\ref{equation: head_additivity}. Furthermore, we can also decompose the input to this attention module at each of the other tokens into the sum of all of its lower-level module outputs in Equation~\ref{equation: residual_decompose}. In Equation~\ref{equation: dictionary_decompose}, since we train a dictionary defined in Equation~\ref{equation: Dictionary_decomposition} on each module output, dictionary learning yields a simple linear decomposition of them.

\begin{align}
    \textbf{Out}_{\text{LXA},i} &= \sum_h \sum_j \textbf{OV}^h_{i \gets j} \nonumber \\
    &= \sum_h \sum_j \textbf{AttnPattern}^h_{ij}W_O^hW^h_Vx_j \label{equation: head_additivity} \\
    &= \sum_h \sum_j \textbf{AttnPattern}^h_{ij}W_O^hW^h_V \sum_{m \in\text{LXA's lower-level modules}}\textbf{Out}_m \label{equation: residual_decompose} \\
    &\approx \sum_h \sum_j \textbf{AttnPattern}^h_{ij}W_O^hW^h_V \sum_{m \in\text{LXA's lower-level modules}}(\sum_{k\in {\text{Dict }m}} {w^m_k \mathbf{d}^m_k}) \label{equation: dictionary_decompose}
\end{align}

The dictionary encoder of LXA output takes in $\textbf{Out}_{\text{LXA},i}$ and applies a linear map $\{ W^{\text{LXA}}_{e, Y}, \mathbf{b}^{\text{LXA}}_{e, Y}\}$ for the pre-ReLU strength of feature Y:

\begin{equation}
    \tilde{w}_{\text{LXAY}, i} \approx W^{\text{LXA}}_{e, Y} \sum_h \sum_j \textbf{AttnPattern}^h_{ij}W_O^hW^h_V\sum_{m \in\text{LXA's lower-level modules}}(\sum_{k\in {\text{Dict }m}} {w^m_k \mathbf{d}^m_k}) + \mathbf{b}^{\text{LXA}}_{e, Y}
\end{equation}

By linearizing the LayerNorm modules (see Appendix~\ref{Appendix: sec: LayerNorm} for more details), we manage to attribute the activation magnitude of Y-th dictionary feature in the i-th token to all dictionary features in the bottom of LXA of all tokens. Such massive linear structure has mostly been discussed in \citet{Elhage2021mathematical}, we further apply it into our framework in the context of dictionary learning.

\subsection{Attributing Attention Patterns to Dictionary Feature Pairs}
\label{section: QK_decomposition}

\begin{figure}[h!]
    \centering
    \includegraphics[width=\linewidth]{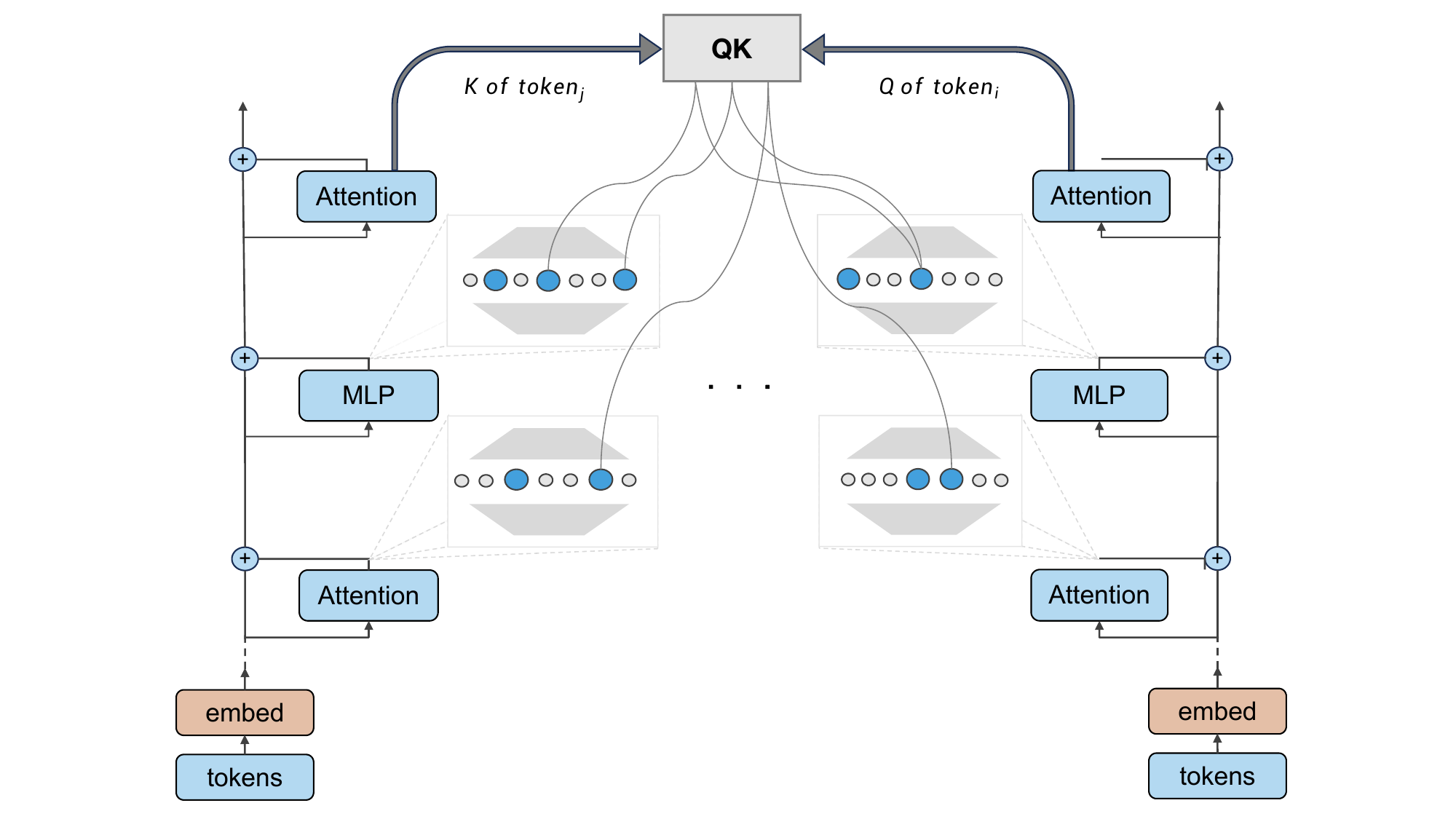}
    \caption{Attention scores (before softmax) decomposed into QK compositions of dictionary feature pairs between the residual stream of two tokens.}
    \label{fig:QK_decomposition}
\end{figure}

In the attention layer of the $X$-$th$ layer i.e. LXA, each head determines how much proportion of its attention from token $i$ be assigned to token $j$ through the following process:

\begin{equation}
    \textbf{AttnPattern}_{ij} = \text{Softmax}(x_iW_QW^\mathrm{T}_Kx^\mathrm{T})_j
\end{equation}

$x\in \mathrm{R}^{L \times D}$ is the input of all tokens to LXA, $x_i\in \mathrm{R}^{1 \times D}$ is input of token $i$ to LXA.
$W_Q, W_K \in \mathrm{R}^{D \times d}$ is the QK circuit of the given attention head. $D, d$ stands for the hidden dimension of the model and the head, respectively.
We omit the superscript indicating head index since QK circuit is independent in each head of LXA.
Throughout this section, we call $x_iW_QW^\mathrm{T}_Kx^\mathrm{T} \textbf{AttnScore}$, and $\textbf{AttnPattern}$ is row-wise normalized $\textbf{AttnScore}$\footnote{This denotation is mainly inspired by Transformer\_lens package \citep{Nanda2022transformerlens}.}.

The input to LXA $x$ is used in both OV and QK circuit of LXA. Thus we can decompose x into linear combination of lower-level dictionary features likewise in the last section, as shown in Equation~\ref{equation: QK_decomposition}. 
For visual effect, we abbreviate \textit{LXA's lower-level modules} to \textit{Lower Modules} in this equation.

\begin{equation}
    \textbf{AttnScore}_{ij} \approx \underbrace{(\sum_{m \in\text{Lower Modules}}(\sum_{k\in {\text{Dict }m}} {c^m_k \mathbf{d}^m_k}))}_{\text{Residual Stream of the } i\text{-th token}}W_QW^\mathrm{T}_K\underbrace{(\sum_{n \in\text{Lower Modules}}(\sum_{k\in {\text{Dict }n}} {c^n_k \mathbf{d}^n_k}))^\mathrm{T}}_{\text{Residual Stream of the } j\text{-th token}}
    \label{equation: QK_decomposition}
\end{equation}

Thanks to the bilinear nature of the QK circuit, we manage to decompose the $\textbf{AttnScore}$ of a single token into contributions of token pairs. Such procedure helps find interpretable \textit{resonance} between two residual streams with respect to dictionary feature pairs.

\subsection{Attributing MLP Features to Lower-level Features}

The MLP module accounts for over half of the parameters in Transformer models, yet research on interpreting MLPs is less than for attention modules in the field of Mechanistic Interpretability.
In the context of dictionary learning, we are able to ask a more specific interpretability question: \textbf{Which subset of lower-level features cause an MLP dictionary feature to activate?}

To answer this, we again start from the output of the MLP module of the $X$-th layer i.e. LXM. It can be simply written as:

\begin{align}
    \textbf{Out}_{\text{LXM}} &= \textbf{LXM}(x_\text{LXM}) \nonumber \\
    &\approx \textbf{LXM}(\sum_{m \in\text{LXM's lower-level modules}}(\sum_{k\in {\text{Dict }m}} {c^m_k \mathbf{d}^m_k})) \label{equation: MLP_decomposition}
\end{align}

The $Y$-th dictionary feature in LXM i.e. LXMY, has the pre-ReLU activation denoted by $\tilde{w}_{\text{LXMY}}$:

\begin{equation}
    \tilde{w}_{\text{LXMY}} = W^{\text{LXM}}_{e, Y} \textbf{LXM}(\sum_{m \in\text{LXM's lower-level modules}}(\sum_{k\in {\text{Dict }m}} {c^m_k \mathbf{d}^m_k})) + \mathbf{b}^{\text{LXM}}_{e, Y}
\end{equation}

\begin{figure}[h!]
    \centering
    \includegraphics[width=.7\textwidth]{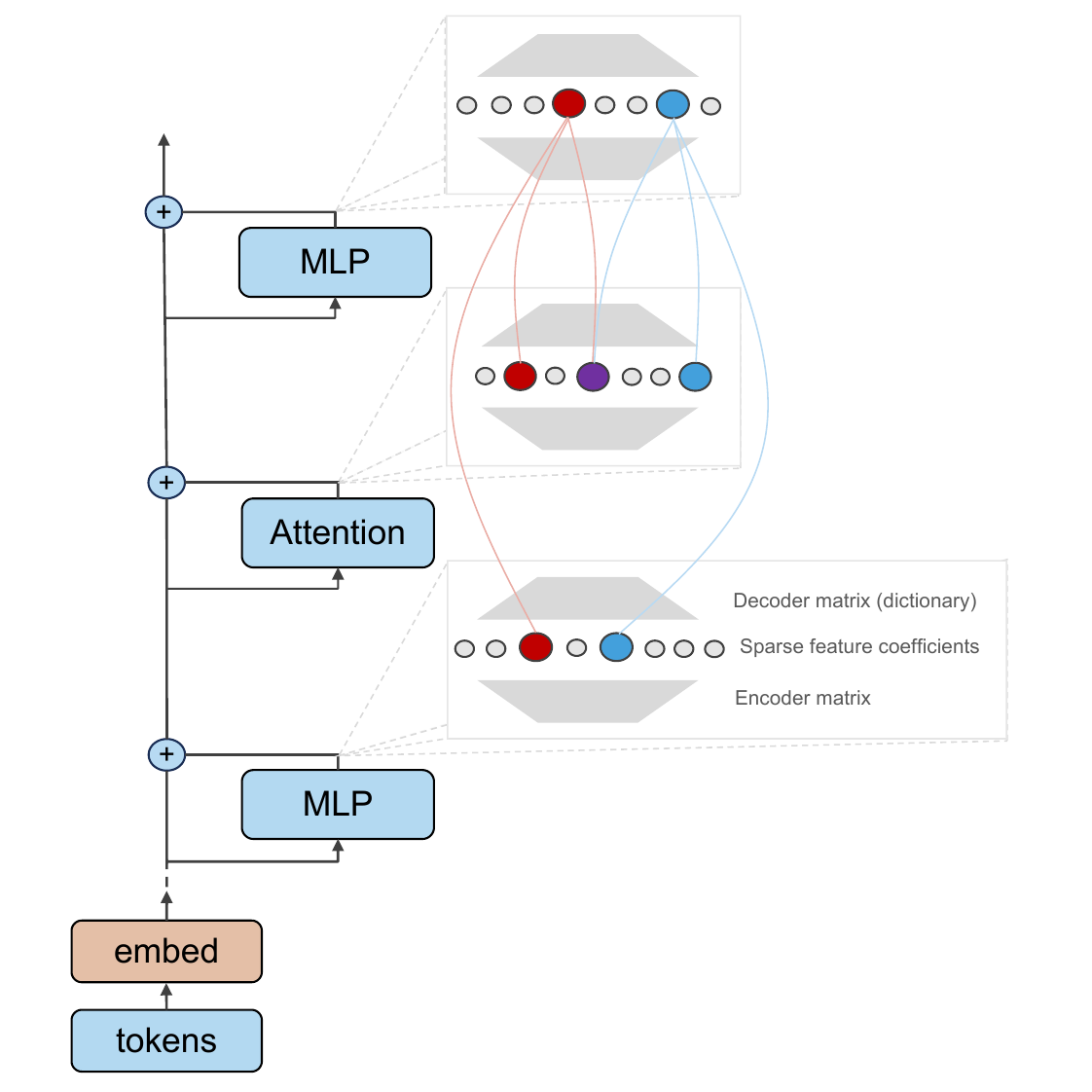}
    \caption{With \textit{Approximate Direct Contribution}, the activation of an MLP feature is decomposed into Approximate Direct Contributions of lower-level dictionary features.}
    \label{fig: MLP_decomposition}
\end{figure}

Unfortunately, MLPs do not share the massive linear nature as attention modules. The non-linearity introduced by the activation function is a central problem, preventing us to directly identify the core contributors of $\tilde{w}_{\text{LXMY}}$. To address this, we introduce \textit{Approximate Direct Contribution} to attribute $\tilde{w}_{\text{LXMY}}$ to each lower-level dictionary feature.

\begin{definition}[Approximate Direct Contribution]
    For an MLP with any self-gated activation function e.g. ReLU\citep{Agarap2018relu}, GeLU\citep{Hendrycks2016GELU} and SiLU\citep{Elfwing2018silu} in the form of $x\cdot\sigma(x)$, the Approximate Direct Contribution \textbf{ADC} of each lower-level dictionary feature $c_k \mathbf{d}_k$ to the activation of LXMY $\tilde{w}_{\text{LXMY}}$ is defined as

    \begin{equation}
    \label{equation: Approximate Direct Contribution}
        \textbf{ADC}(c_k \mathbf{d}_k) = W^{\text{LXM}}_{e, Y} W_{out}(\underbrace{(W_{in}c_k \mathbf{d}_k)}_{\text{Dictionary Feature}}\cdot\overbrace{\sigma(W_{in}x)}^{\text{Leave MLP input unchanged for }\sigma})
    \end{equation}
\end{definition}

The intuition behind $\textbf{ADC}$ is quite simple. The non-linear MLP is transformed into a linear function of contributions of input features. We treat the input of the activation function as a constant, and the other item $W_{in}c_k \mathbf{d}_k$ as a linear function of the input features.

\subsection{Circuit Discovery}

We have introduced how we tackle with QK, OV and MLP in the context of dictionary learning. By understanding the composition of each monosemantic dictionary feature and iteratively attributing features in a top-down manner, we are theoretically able to understand most information flow through the residual stream and across residual streams.

Our theoretical framework encompasses discovering both end-to-end and local circuits. We can start from the output logit or any dictionary feature and end with any feature or the input embedding.

In terms of circuit granularity, contributions to attention outputs via the OV circuit can be linearly assigned to each head. And \textit{Approximate Direct Contribution} is also a linear function of the activation of each single neuron.

\section{Dissecting Othello-GPT}

\subsection{Background and Notations}

We followed \citet{Li2022othello} and trained a 1.2M parameter decoder-only Transformer to learn a programmable game prediction task named Othello.
The model only learns to play legal moves, not tactics.
The rule of Othello is as follows:
Two players compete, using 64 identical game pieces that are light on one side and dark on the other. Each player chooses one color to use throughout the game. Players take turns placing one disk on an empty tile, with their assigned color facing up. 
After a play is made, any disks of the opponent's color that lie in a straight line bounded by the one just played and another one in the current player's color are turned over. The game ends till there are no more valid moves for both players.

Our Transformer model consists of 6 layers with a hidden dimension of 128. For more details of model and dictionary training please refer to Appendix~\ref{Appendix: Details}. We follow the denotation in the previous section. The $Y$-th dictionary feature in the $X$-th layer's attention/MLP layer is denoted as LXAY/LXMY.

The tiles are numbered from 1 to 64 and the model learns auto-regressively to predict the next valid move from a sequence representing a synthetic game. Prior work\citep{Li2022othello, Hazineh2023OthelloLinear, Nanda2023emergent} has found surprising linear structure in Othello-GPT by probing. We recommend readers to see Section~\ref{section: Othello-GPT} for more details. Though the model's input is integers ranging from 1 to 64, a more intuitive notation is labeling rows from a to g and columns 1 to 8. For instance, the 9th pile corresponds to b-1.

\subsection{Dictionary Features in Othello-GPT}

Since we have trained dictionaries on each module's output as mentioned in Section~\ref{section: where to train?}, we are able to tell what information each module write into the residual stream.

Here is a brief summary of dictionary features we have found:

\begin{itemize}
    \item \textbf{Current move position features:} There are many features in L0A and L0M that respond to the current move being at a specific position, e.g. L0A53: current move at g-3, L0A80: current move at e-0, L0M996: current move at a-4 flipping the piece above, L0M195: current move at c-1 flipping the piece on the right, L0M205: current move at c-1 flipping the piece at top right. Such features are also present in L1M, L2M and L3M.

    \item \textbf{Features representing board state:} There are many features in L1A to L4M i.e. middle layers describing the current board state, e.g. L1A626: f-1 is opponent's piece, L1M158: e-2 is own piece. 

    \item \textbf{Features for empty cells:} L5A contains many features describing a cell being empty, e.g. L5A336: c-3 is empty.

    \item \textbf{Features indicating legal moves:} L5M contains many features indicating a position is a legal move, e.g. L5M733: b-3 is legal. These features also directly contribute to their corresponding logit when activated.
\end{itemize}

Dictionary learning finds \textit{all} types of features found in previous work\citep{Nanda2023emergent, Hazineh2023OthelloLinear}. 

Please refer to Appendix~\ref{Appendix: Interface} for more details of our interpretation interface.

\subsection{A Case Study on Circuits in Othello-GPT}

In this section, we will showcase circuits regarding some certain types of dictionary features. An important clarification is that our circuit analysis is based on certain inputs and we claim no universal understanding of Othello-GPT. Discovering general motif in Transformers is a central problem in Mechanistic Interpretability and is left for future work.

\begin{figure}[h!]
     \centering
     \begin{subfigure}[m]{\textwidth}
         \centering
         \includegraphics[width=.5\textwidth]{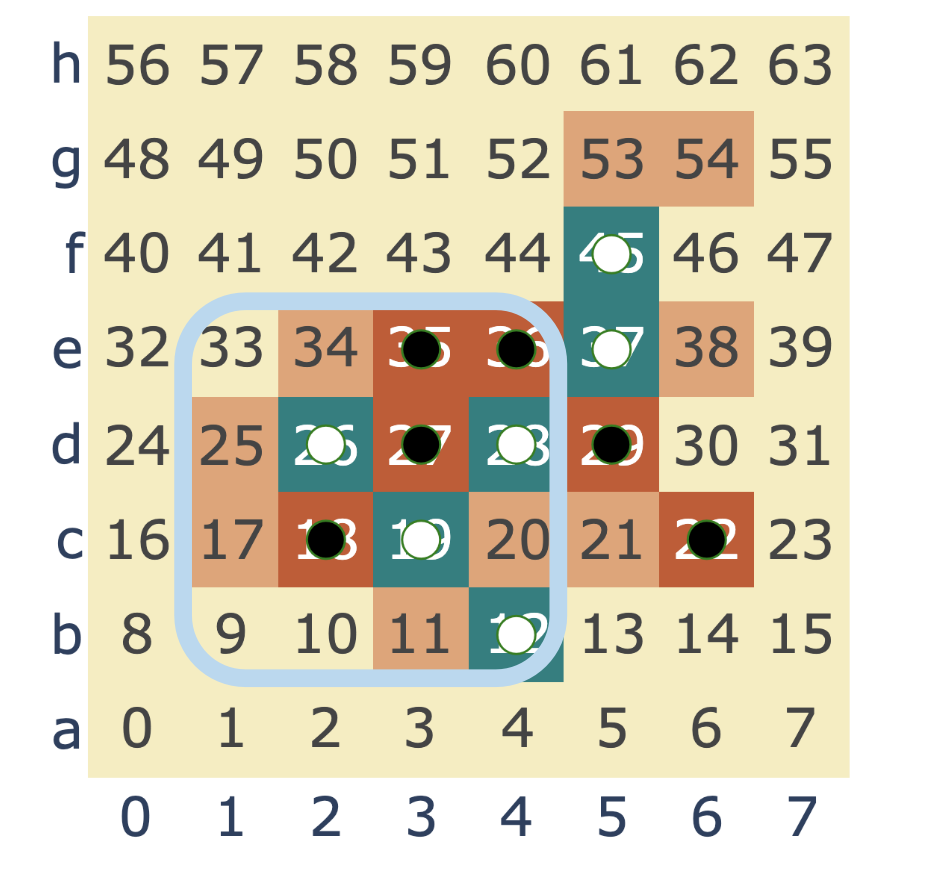}
         \caption{Board state at move 8. White player has just played on b-4 and flipped c-3, so throughout this residual stream Othello-GPT treats white tiles as mine and black as the opponent's. We mainly focus on the azure-framed local state in Figure~\ref{fig: OV_example illustration}.}
         \label{fig: OV_example board state}
     \end{subfigure}
     \hfill
     \begin{subfigure}[m]{\textwidth}
         \centering
         \includegraphics[width=\textwidth]{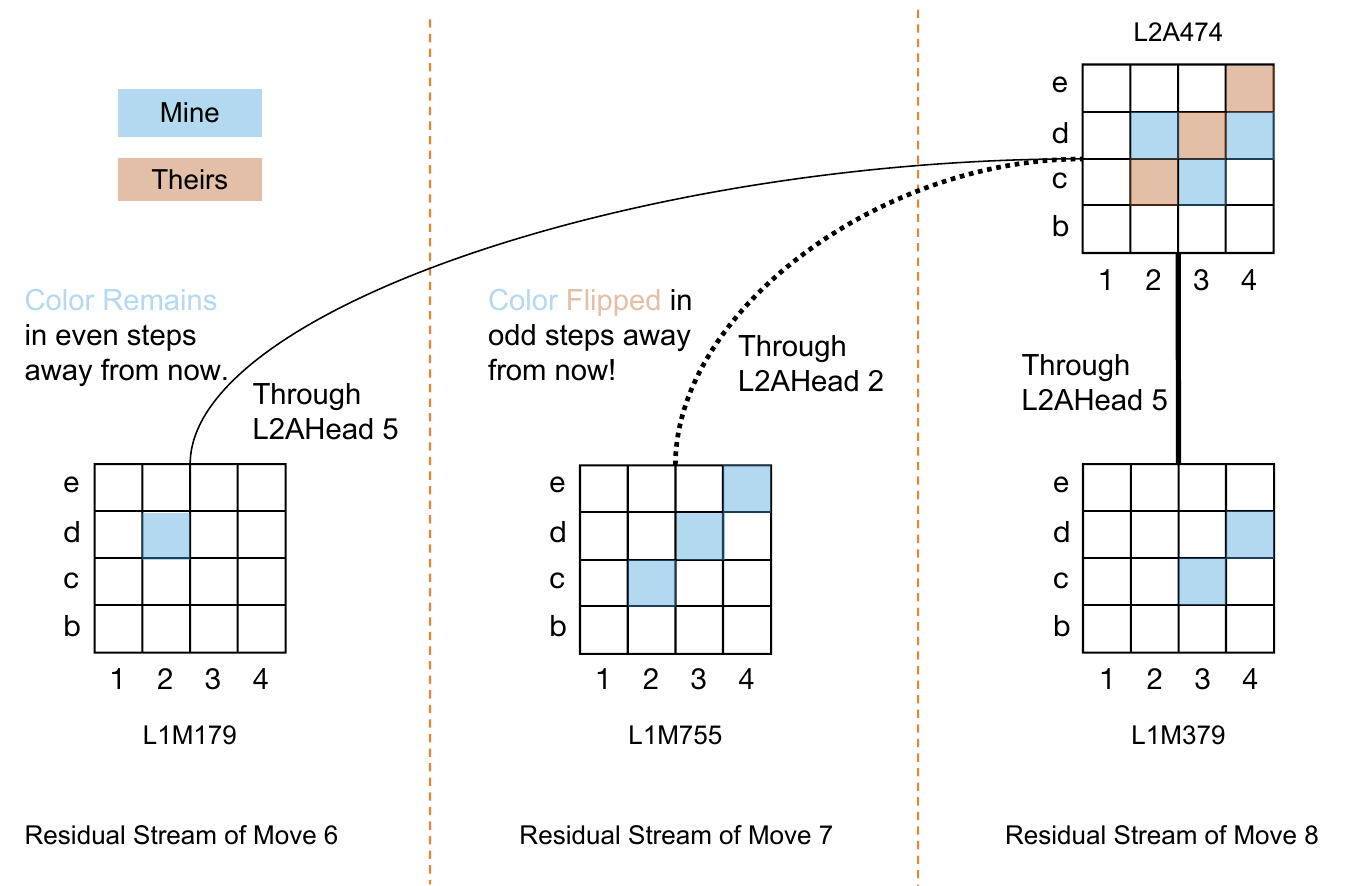}
         \caption{A simplified illustration of how Othello-GPT manage to derive the correct local board state around d-3 in the attention layer. In brief, it combines the board state computed earlier and the updated state in the current residual stream (represented by lower-level L1M dictionary features) via specific attention heads in the OV circuit.}
         \label{fig: OV_example illustration}
     \end{subfigure}
        \caption{An example of interpretable information transition via the OV circuit.}
        \label{fig: OV_example}
\end{figure}

\subsubsection{A Local OV Circuit Computing the Board State}
\label{section: OV_example}

We fed the model with a randomly chosen input and decomposed the dictionary feature with the strongest activation in L2A i.e. L2A474 with our theoretical framework introduced in Section~\ref{section: theory_OV_circuit}. We have found a circuit computing the board state after the current move with information across multiple moves stored in KV, as shown in Figure~\ref{fig: OV_example}.

The top 3 contributors from 3 different residual streams explain the 6 tiles this feature represent. 
Specifically, L1M179 indicates that d-2 was white in move 6 and is viewed as my color in both move 6 and 8 because they are both white moves. So the OV circuit copies this information to L2A and thus makes contribution to L2A474 which also tends to activate when d-2 is mine. 
The 3 tiles from c-2 to e-4 was black (mine) in move 7 so Othello-GPT rather treats them as the opponent's in move 8. These two facts are represented by L1M755 in the residual stream of move 7 and L2A474 in move 8, correspondingly.
By L1M in move 8, we can see Othello-GPT has realized that c-3 has been flipped so it views c-3 as mine, indicated by L1M379. Then it copies this information to L2A474 via the OV circuit.

In this example, we fail to observe significant attention superposition\citep{Larson2023ExpandingSuperposition,Greenspan2023AttentionSuperposition}. Both L1M379 in move 8 and L1M179 in move 6 contributes to L2A474 in move 8 through head 5 in L2A. However, retrieving information from the opponent's move utilizes head 2. From left to right in Figure~\ref{fig: OV_example illustration}, their corresponding most important head accounts for 110\%, 105\% and 113\% of the overall contribution through L2A OV circuit to L2A474, respectively. And other heads under each circumstance all have near zero contribution.
We speculate that L2A head 5 attends to \textit{my} moves and head 2 attends to \textit{the opponent's} moves, which can be inferred from their attention patterns.

Actually, the top 3 contributors of L2A474 only accounts for a small proportion of its activation magnitude. Nonetheless, most other contributors share the same or near semantics because of redundancy of features, even if we did not include dropout during training.

\subsubsection{How Do QK Circuits Implement 'Attending to Theirs' Heads?}

\begin{figure}[h!]
     \centering
     \begin{subfigure}[b]{.3\textwidth}
         \centering
         \includegraphics[width=\textwidth]{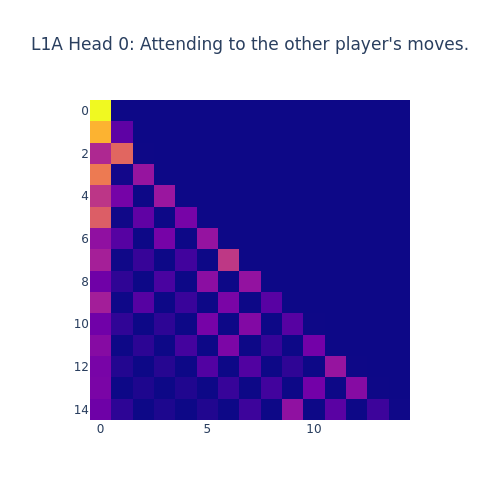}
         \caption{Head 0 of L1A attends to the opponent's moves.}
         \label{fig: QK_example_attn_pattern}
     \end{subfigure}
     \hfill
     \begin{subfigure}[b]{.6\textwidth}
         \centering
         \includegraphics[width=\textwidth]{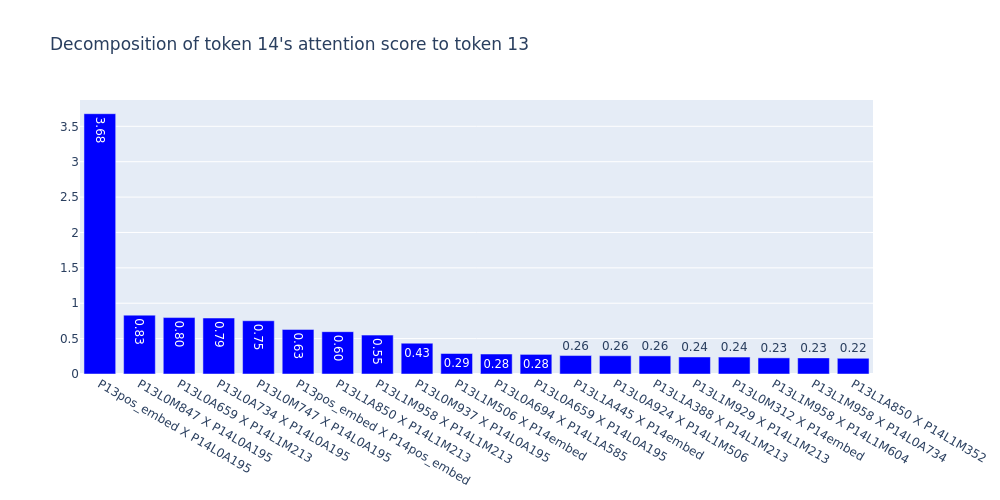}
         \caption{Feature pairs contributing to the attention score from token 14 to token 13.}
         \label{fig: QK_example_decomposition}
     \end{subfigure}
        \caption{An example of interpretable decomposition of QK attention scores.}
        \label{fig: QK_example}
\end{figure}

Another important question to ask is how Othello-GPT implement its two main kinds of attention pattern i.e. attending to my and the opponent's move, as observed in \citet{Nanda2023emergent}.

We showcase the most important feature pairs contributing to a salient attention score in L1A head 0. We omit the specific input in this section since this is a relatively universal behavior for L1A head 0 to attend to the opponent's moves and form a attention pattern like shown in Figure~\ref{fig: QK_example_attn_pattern}.

The bilinear QK circuit can be decomposed into \textit{resonance} between feature pairs as described in Section~\ref{section: QK_decomposition}, and the result is shown in Figure~\ref{fig: QK_example_decomposition}. The most important features in token 13 e.g. position embedding, L0M847 and L0A659 all indicates that token 13 is white. Similarly, L0A195 and position embedding of token 14 means this is a black move. Thus we conclude position embedding and low-level features representing the color of the current move as a whole lead to the attention pattern as shown in Figure~\ref{fig: QK_example_attn_pattern}.

\subsubsection{How Do Early MLPs Identify Filpped Pieces?}

In the example illustrated by Figure~\ref{fig: OV_example board state} in Section~\ref{section: OV_example}, one intriguing fact is that Othello-GPT has realized that c-3 has been flipped from black to white by L1M since L1M379 indicates that both c-3 and d-4 is \textit{my} color i.e. white. We believe the mechanism behind this is not trivial in that filpping a piece requires combining the state of at least 3 tiles in a line. Circuit analysis with \textit{Approximate Direct Contribution} reveals a clean structure of how L1M recognize c-3 being flipped in this case.

\begin{figure}[h!]
     \centering
     \begin{subfigure}[m]{\textwidth}
         \centering
         \includegraphics[width=.7\linewidth]{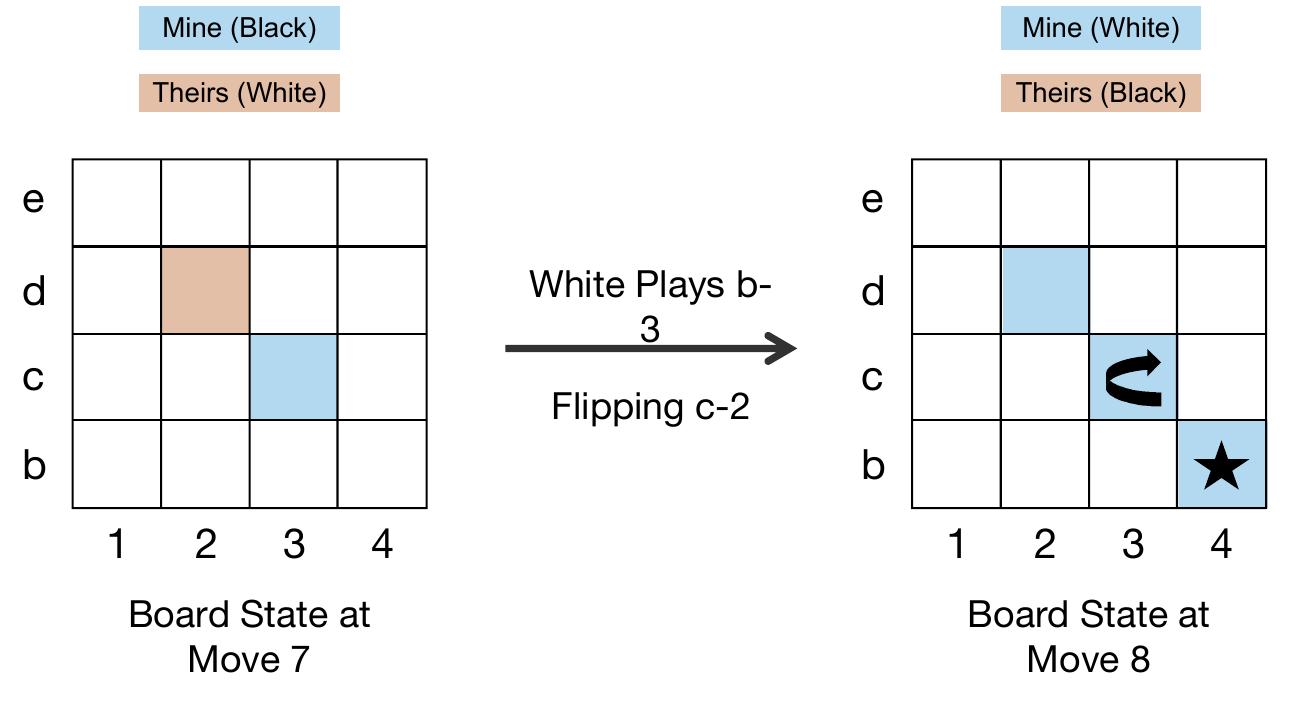}
         \caption{In move 8, white plays at b-3 and flipped c-2 together with the white piece at d-1.}
         \label{fig: MLP_example_flip}
     \end{subfigure}
     \hfill
     \begin{subfigure}[m]{\textwidth}
         \centering
         \includegraphics[width=\textwidth]{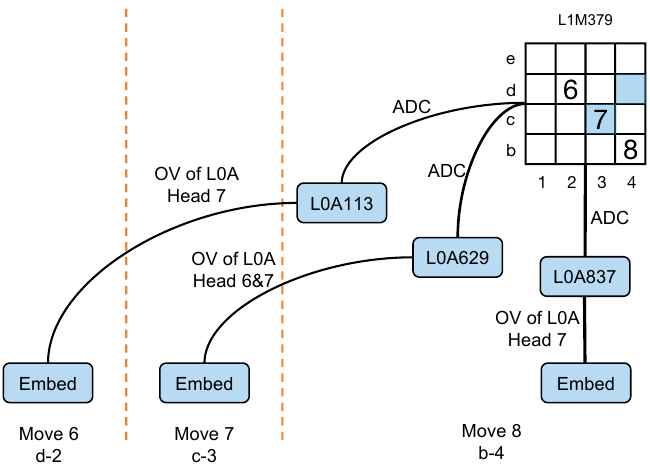}
         \caption{\textit{Approximate Direct Contribution} decomposition identifies 3 L0A dictionary features, which respectively retrieve the piece played at move 6, 7 and 8. OV circuit adds \textit{my} or \textit{the opponent's} color attribution then L1M serves as a \textit{and} gate to derive c-3 being flipped.}
         \label{fig: MLP_example_illustration}
     \end{subfigure}
        \caption{An example of interpretable decomposition of MLP dictionary features.}
        \label{fig: MLP_example}
\end{figure}

Figure~\ref{fig: MLP_example_flip} illustrates the local state transition that c-2 is flipped from black to white. Our circuit analysis regarding the formation of L1M379 are shown in Figure~\ref{fig: MLP_example}. The three lower-level features with the largest \textit{Approximate Direct Contribution} correspond to three consecutive moves played on a line. However, Othello-GPT still needs to recognize a white-black-white structure to confidently flip the black piece in the middle in its internal. From the observation in Section~\ref{section: OV_example}, we conclude it is specific heads of L0A that introduces the information of piece color. A strong evidence is that OV decomposition in the granularity of attention heads shows that both head 6 and 7 contributes to L0A629 while the other two features are only fired by head 7. This also implies that we succeed to observe attention superposition i.e. head 6 and 7 of L0A collaborating to activate L0A629 in the wild.

Finally, L1M makes an \textit{and} operation on these features and yields the conclusion that c-3 has been flipped to \textit{my} color.

\section{Related Work}

\subsection{Mechanistic Interpretability}

Mechanistic Interpretability\citep{Olah2020zoom} seeks to fully reverse engineer a trained neural network. We mainly focus on Transformer-based language models in this section. A speculative approach to achieving this ultimate goal is to: 1. Attacking superposition\citep{Arora2018LinearStructure,Elhage2022superposition} to decompose neural activations into human-understandable primitives. 2. Studying the learned parameters rigorously to understand how neural networks compose lower-level features into higher-level ones, and how they finally use these features to make decisions.

Existing literature in Mechanistic Interpretability roughly follows a looser workflow. Researchers define a behavior that the model exhibits e.g. modular arithmetic\citep{Zhong2023clock,Nanda2023clock}, Indirect Object Identification\citep{Wang2023IOI}, factual retrieval\citep{Meng2022ROME} or more abstract behaviors like in-context learning\citep{Olsson2022induction} and Chain-of-Thought reasoning\citep{Anonymous2024MICoT}. Activation patching and direct logit attribution are two powerful tools to locate factual knowledge or identify key attention heads implementing a certain \textbf{end-to-end} behavior.

\subsection{Attacking Superposition and Extracting More Monosemantic Features}

One critical reason why little work is done on language models to find local circuits is the superposition hypothesis\citep{Elhage2022superposition} . It assumes that the model learns to cram more linear features than it has hidden dimensions into the activation space, making it problematic to interpret the hidden activations. Additionally, the sparse nature of features in language models make it more challenging to find monosemantic neurons than in vision models\citep{Cammarata2020curve,Goh2021multimodal}.

There are two main approaches to attacking superposition, namely architectural and mechanistic one. The former encompasses eliminating superposition with inductive bias in the training phase\citep{Elhage2022solu,Yu2023WhiteBoxTRM}. This work lies in the latter realm. Sparse dictionary learning aims to train a sparse autoencoder with sparsity constrain on the hidden layer. Prior work managed to decompose word embedding\citep{Yun2021dict, Chen2017dict, Zhang2019dict, Panigrahi2019dict}, MLP hidden layer\citep{Bricken2023monosemanticity} and the residual stream\citep{Cunningham2023dict} into more interpretable linear features.

\subsection{Circuit Discovery}

\citet{Conmy2023ACDC} systemized the workflow of discovering circuits in existing works. They first define the granularity of circuit analysis e.g. attention head/layer and MLP neuron/layer and patching each basic component to quantify its importance via its effect on the final logit of an interested token. The computational graph can be then pruned to a subgraph representing the key components implementing an algorithm. Such method succeeded in finding many circuits in various size of language models\citep{Wang2023IOI,Anonymous2024MICoT,Heimersheim2023docstring,Lieberum2023ChinchillaCircuit}.

To solve the out-of-distribution problem, \citet{Zhang2023Bestpractice} discussed the best practice to mitigate distribution shift. It has also been pointed out that patching-based methods yield some counter-intuitive results like the hydra effect\citep{McGrath2023Hydra} (also known as backup behavior in \citet{Wang2023IOI}).

To our best knowledge, little work has been done to find a patch-free counterpart to discover circuits in a patch-free manner despite the aforementioned problems.

\subsection{Understanding Othello-GPT}
\label{section: Othello-GPT}

The task of next move prediction in Othello game was originally proposed in \citet{Li2022othello}. 
The main idea of the original work was to validate the formation of world models on this task.
Through \textbf{non-linear probing}, the authors provided strong evidence for world model formation: the model first computes the state of the board and makes decisions accordingly, rather than simply memorizing surface patterns.
\citet{Nanda2023emergent} and \citet{Hazineh2023OthelloLinear} then found that the conclusions in the original paper were insufficient. 
If the probe target is changed from (black/white/empty) to (my color/the opponent's color/empty), \textbf{linear probes} can also achieve good performance, yielding a strong evidence of the linearity assumption of features\citep{Mikolov2013w2v}.

\section{Discussions}

\subsection{Comparison to Patch-Based Methods}

This section encompasses two aspects of comparison between our theoretical framework and existing patch-based methods. Overall, we do not claim overwhelming superiority over patching dictionary features for circuit discovery. We recommend readers to view our method as an efficient alternative to feature ablation.

\subsubsection{Dimensions of Choice for Feature Patching}

In the context of dictionary learning, there are two dimensions of choice for patching:

\begin{itemize}
    \item \textbf{Zero, mean or less-than-rank-one?} What value should we reset our interested dictionary feature to? As discussed in \citet{Zhang2023Bestpractice}, in existing workflow of activation patching, mean patching i.e. replacing one activation with the average activation of several random inputs is preferred to zero patching i.e. with zeros. However, since dictionary features are usually sparse\citep{Elhage2022superposition,Bricken2023monosemanticity}, these two choices appear to have no significant difference under most circumstances.

    Another choice adopted in \citet{Cunningham2023dict} is less-than-rank-one ablation. Instead of modifying the dictionary decomposition, they subtract from the original activation its projection on the encoder row of the interested feature, which also affects other features. Since their configuration in dictionary training is divergent from ours, we do not include this choice in later discussion.

    \item \textbf{Causal or direct?} We call it \textbf{causal ablation} if one low-level feature is ablated and the corrupted activation is fed into subsequent modules for re-computation till we see the effect on our interested feature. Another choice is to directly linearize all low-level dictionary features except one through the residual stream to the interested module and study the consequent activation or logit difference. We call the latter \textbf{direct ablation}.
\end{itemize}

\subsubsection{Computational Complexity}

\begin{table}[h!]
\centering
\begin{tabular}{@{}ccc@{}}
\toprule
\textbf{Methods}         & \textbf{Asymptotic Complexity} & \textbf{Out-of-Distribution} \\ \midrule
\textbf{Causal Patching} & O(n) model forwards            & Yes                          \\
\textbf{Direct Patching} & O(n) module forwards           & Yes                          \\
\textbf{Ours}            & O(n) module forwards           & No                           \\ \bottomrule
\end{tabular}
\caption{To study how each lower-level feature contributes to a high-level feature, causal patching requires number of model inference in proportion to the number of lower-level features $n$. Direct patching and our method only need to re-run the interested module. Patch-based methods face the problem of out-of-distribution which our framework do not suffer from.}
\label{table: complexity}
\end{table}

As the model and dictionary we study go larger, the number of lower-level features grows correspondingly. Table~\ref{table: complexity} shows the comparison between our method and patching-based methods. Our framework achieves the optimal complexity and is free from out-of-distribution.

\subsubsection{Correspondence between Approximate Direct Contribution and Direct Patching}

We mainly focus on comparison between \textit{Approximate Direct Contribution} and direct patching in attributing MLP features since our analysis for OV and QK circuit are based on linear structures and should theoretically achieve the same result as direct patching.

We randomly sample 10 input sequences and get the top 3 most activated dictionary features in all 6 MLP layers i.e. 180 features. We find the top 5 most important lower-level features for each MLP feature with both \textit{Approximate Direct Contribution} and direct patching and computed the intersection over union of their corresponding contributors. The average IOU is 0.68, indicating a relatively high correspondence of these two methods. This can also be viewed as a evaluation of using \textit{Approximate Direct Contribution} to find which feature activated an interested MLP feature.

\subsection{Limitations of QK Decomposition and Approximate Direct Contribution}

\subsubsection{QK Decomposition}

As an important non-linear module in Transformers, Softmax normalizes the attention score, and this processing causes any change in the elements of the attention scores to affect the attention pattern across the entire sequence, making it difficult for explanation. Our QK decomposition procedure only includes decomposing each single attention score and fails to interpret their relative strength. We believe more work needs to be done to gain more comprehensive understanding of the attention pattern.

\subsubsection{Approximate Direct Contribution}

\textit{Approximate Direct Contribution} of the whole MLP module, as defined in Equation~\ref{equation: Approximate Direct Contribution}, equals to the sum of each neuron's \textbf{ADC} since MLP has a privileged basis defined by neurons\citep{Elhage2023privileged}. Therefore, we discuss the limitation of \textbf{ADC} with respect to a single neuron.
For an MLP with a self-gated activation function in the form of $x\cdot\sigma(x)$, if we consider the contribution of a single input feature $c_k \textbf{d}_k$ to the output feature $\tilde{w}_{\text{LXMY}}$, there are only two sources of contribution: neurons with a positive (linear) effect on $\tilde{w}_{\text{LXMY}}$ are activated by $c_k \textbf{d}_k$ and neurons having a negative (linear) effect on $\tilde{w}_{\text{LXMY}}$ are inhibited by $c_k \textbf{d}_k$.

\textit{Approximate Direct Contribution} can qualitatively capture the former type of contribution since for any monotonically non-decreasing non-negative function $\sigma(x)$, $\frac{\partial}{\partial x}  x\sigma(x)$ is always non-decreasing.
Specifically, if an input feature $c_k \textbf{d}_k$ adds on an MLP neuron, and that neuron is activated via $\sigma(x)$ and contributes to some output feature $\tilde{w}_{\text{LXMY}}$, then this neuron propagates a positive \textbf{ADC} from $c_k \textbf{d}_k$ to $\tilde{w}_{\text{LXMY}}$.

However, the latter type of contribution, suppressing neurons with negative effects on $\tilde{w}_{\text{LXMY}}$, is not captured well by \textbf{ADC}. Since we fix $\sigma(x)$ as constant, the real contribution will be reflected non-qualitatively in \textbf{ADC} if and only if $\sigma(x)$ is not inhibited to 0. Otherwise, \textbf{ADC} through this neuron will be 0.

The two aforementioned kinds of contribution is not symmetric, however. Without contributions of the first type, any dictionary feature can only be activated up to a certain limit i.e. by bias terms like decoder biases of dictionaries or layer norm biases. Empirically, we find that most MLP features can be decomposed interpretably via \textbf{ADC} analysis, in which decoder biases has little contribution.

\begin{figure}[h!]
    \centering
    \includegraphics[width=\linewidth]{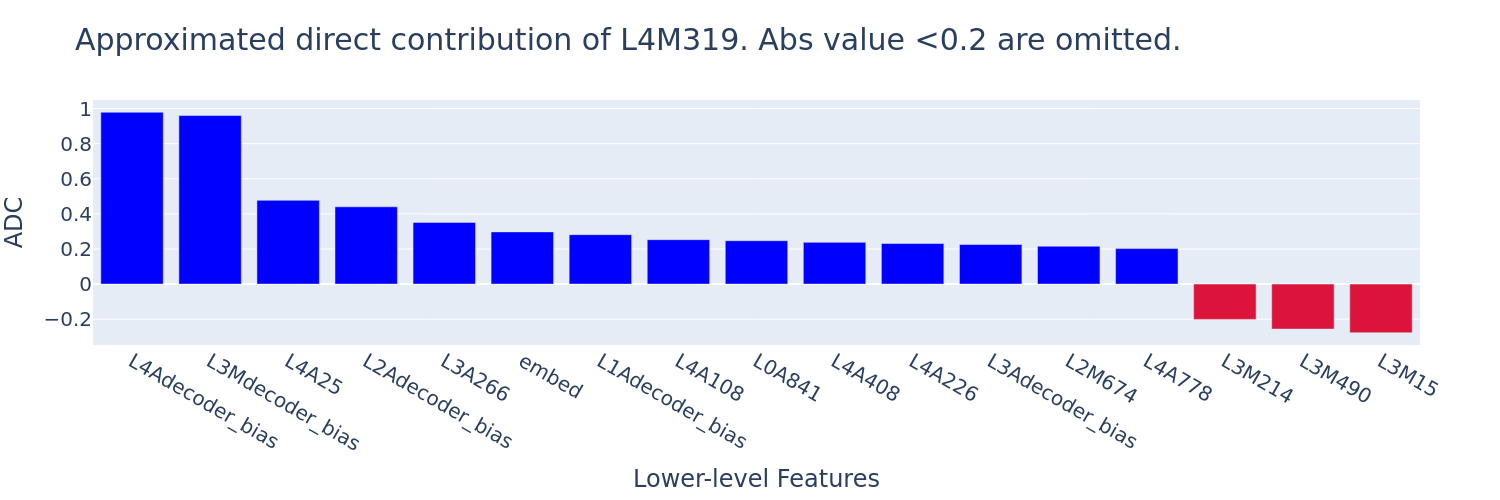}
    \caption{\textit{Approximate Direct Contribution} attributes the activation of }
    \label{fig: MLP_failure_case}
\end{figure}

There are also failure cases as shown in Figure~\ref{fig: MLP_failure_case} where \textbf{ADC} analysis identifies L4A decoder bias and L3M decoder bias as the main contributor to L4M319. We conjecture that it is due to an MLP neuron implementing an \textit{or gate} as discussed in \citet{Conmy2023ACDC} in the form of $c = 1 - \text{ReLU}(1 - a - b)$. In this case, \textbf{ADC} will attribute to a constant rather than dictionary features. It is worth mentioning that both our method and patch-based method theoretically cannot identify both feature in an \textit{or gate}.

\subsection{Future Work and Open Problems}

This work lies in two new and rising fields of dictionary learning and Othello interpretability. There remain a large number of problems we have encountered. 

One thing we are most excited about is \textit{Transformer Pathology} i.e. identifying the features resulting in undesired outputs. One speculative motivation for these pathogenic features is imperfect portfolio of opposite features. For example, OV circuit may inevitably retrieve outdated board states from earlier moves, for which is often compensated by later states or flipped features in the current residual stream. Such mechanism sometimes fail to balance and wrong board states take the lead, causing wrong decisions. Due to our sub-optimal dictionary training procedure, we leave this for future work to present a cleaner structure inside of Othello-GPT.

We are also interested in scalable analysis of circuits. Even in a small Othello model, there exists about 1,000 active dictionary features in each residual stream. An important future work is to find feature families and establish taxonomy of circuits in both Othello-GPT and language models.

\section{Conclusions}

We propose a theoretical framework to discover circuits in Transformers with dictionary learning. Our method requires no activation patching and is computationally efficient. One important theoretical contribution of ours is the discussion of positions to train dictionaries on and we claim it sensible to decompose every module writing to the residual stream, including the input embedding. We test our method on Othello-GPT and discovered a number of interpretable circuits. However, we only present some representative cases with clean structures. We believe better dictionary training techniques and presentation interface may help us share more universal findings with our framework in Othello-GPT. We are also looking forward to applications in real-world language models.

\subsubsection*{Acknowledgments}
The open-source research on Othello-GPT of Kenneth Li\citep{Li2022othello} and Neel Nanda\citep{Nanda2023emergent} offers much inspiration to this work. Our data generation procedure and visualization of board state are heavily based on their code.
Our implementation of activation caching and circuit analysis relies on Transformer\_lens\citep{Nanda2022transformerlens}. We would also like to thank \citet{Cunningham2023dict} and \citet{Bricken2023monosemanticity} for open-sourcing their dictionary training procedure and experimental details, which offered us with significant convenience.

The computing resources used in this work are supported by the Computing for the Future at Fudan (CFFF) group. The enthusiasm and professionalism of CFFF staff provided a great guarantee for the smooth progress of this work.

\bibliography{colm2024_conference}
\bibliographystyle{colm2024_conference}

\appendix
\section{Appendix}

\subsection{Dealing with Non-linearity of LayerNorm}
\label{Appendix: sec: LayerNorm}
In prevalent transformer architectures, LayerNorm is usually applied after each module makes a copy of the residual stream, i.e. pre-norm. 
Although the input to each module can be linearly decomposed into the sum of outputs from all bottom modules, LayerNorm itself is not a linear operation. 
Concretely, the step of calculating the standard deviation is non-linear to the input of LayerNorm.
This prevents us from attributing a certain consequence to each linear component, which is an important issue to resolve for facilitating our circuit analyses.

To address this, we treat the standard deviation of the input as a constant rather than a function of it. 
This allows us to transform LayerNorm into a linear function of the input without changing the computational result.
With this transformation, we can now apply the modified LayerNorm separately to any linear decomposition of the input activation to estimate the impact of each component on the result.

This trick is widely applied in existing mechanistic interpretability work\citep{Wang2023IOI, Heimersheim2023docstring} to attribute the output logit of a Transformer model to the output of a given module i.e. Direct Logit Attribution.

\subsection{Experimental Details}
\label{Appendix: Details}

\begin{figure}[h!]
     \centering
     \begin{subfigure}[m]{\textwidth}
         \centering
         \includegraphics[width=\textwidth]{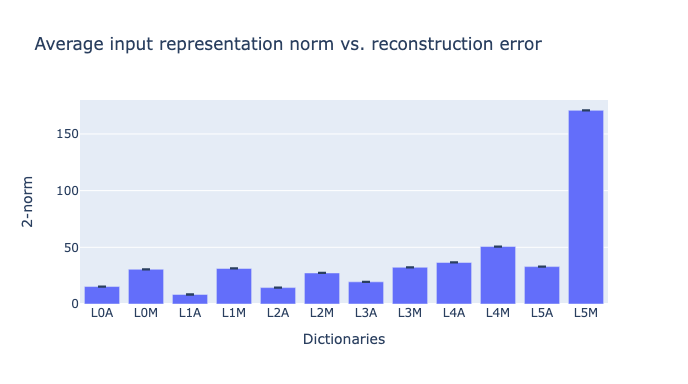}
         \caption{Illustration of the average L2 norm of each module output. The nearly invisible error bar shows the reconstruction error of each dictionary, converted from MSE to L2 norm.}
     \end{subfigure}
     \hfill
     \begin{subfigure}[m]{\textwidth}
         \centering
         \includegraphics[width=\textwidth]{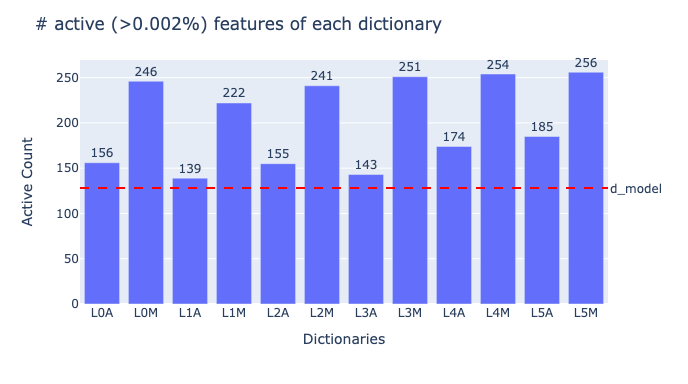}
         \caption{Number of active feature (activated at least once in 50000 inputs) of each dictionary. The hidden layers of the dictionaries consist of 1024 neurons.}
     \end{subfigure}
        \caption{Limitation of our dictionary training: near perfect reconstruction error and too much dead neurons.}
        \label{fig: Dictionary_limitation}
\end{figure}

\subsubsection{Our Re-trained Version of Othello-GPT}

Although the original Othello-GPT paper\citep{Li2022othello} has open-sourced the model architecture and parameters, this article makes some modifications and retrained the model:

The model has about 1.2 million parameters and a hidden dimension of 128. Each MLP hidden layer has 512 neurons. We trained out Othello model on about 1e9 sequences.

\begin{itemize}
    \item \textbf{Removing Dropout:} Dropout naturally aggravates redundancy which increases the difficulty of circuit analysis. This setting is also widely applied in mainstream language model training pipeline\citep{Touvron2023llama2}.

    \item \textbf{Reducing model parameters:} Without sacrificing model performance (99.9\% in the original paper drops to 99.5\%) or interpretability conclusions, we made the model shallower from 8 layers to 6 layers to reduce repetitive experimental sections.
\end{itemize}

\subsubsection{Dictionary Training}

We adopt a vanilla dictionary training pipeline with untied encoder-decoder and L1 sparsity constraint. We trained an autoencoder with 128 neurons in input and output layers and 1024 in the hidden layer. We observe the same trend as in \citet{Dettmers2022LLMint8} that the activation of Transformers goes larger in later layers, for which we empirically use smaller L1 sparsity coefficient $\alpha$.

One limitation in our dictionary training is that we may have used too small $\alpha$'s. In consequence the reconstruction performance is near perfect and we observed about 75\% dead neurons in each dictionary, as shown in Figure~\ref{fig: Dictionary_limitation}.

We believe deeper understanding of Othello-GPT can be achieved given this problem is alleviated.

\subsection{Interface for Interpretability}
\label{Appendix: Interface}

\begin{figure}[h!]
    \centering
    \includegraphics[width=\linewidth]{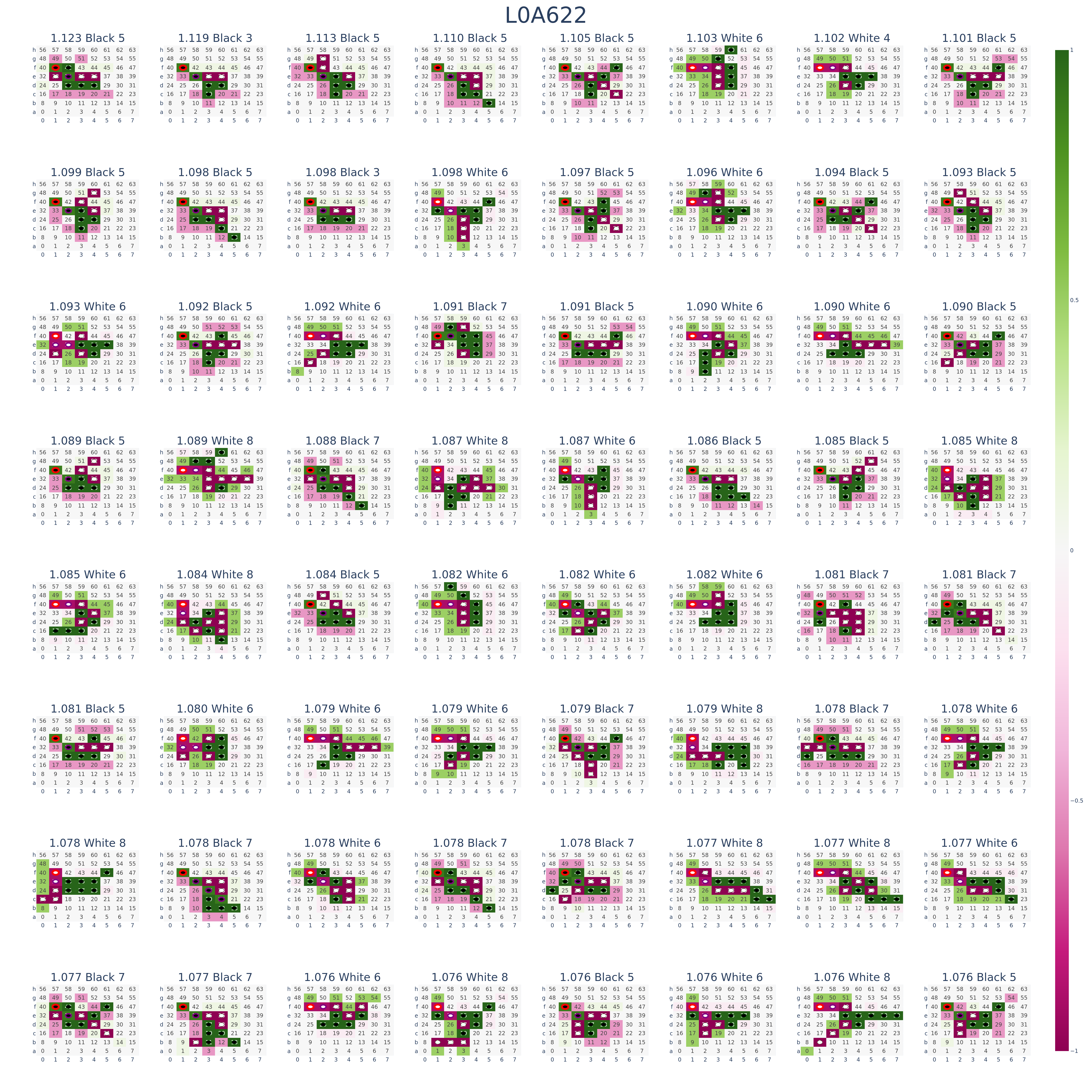}
    \caption{64 board states that activate L0A622 the most.}
    \label{fig: L0A622}
\end{figure}

\begin{figure}[h!]
    \centering
    \includegraphics[width=\linewidth]{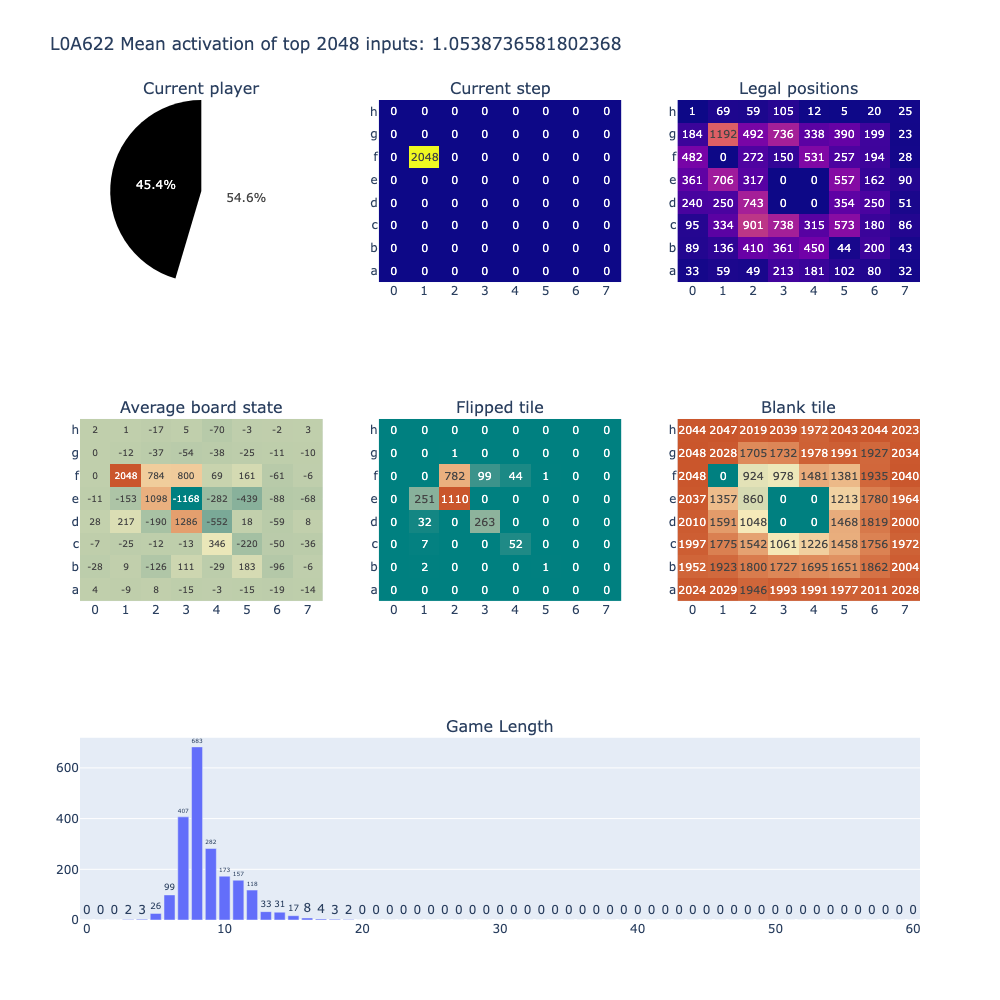}
    \caption{Our interpretability interface for dictionary features. There are 7 plots showing a statistic of 2048 most activating ones out of 1.2 million board states.}
    \label{fig: L0A622_interface}
\end{figure}

For each feature, we examine the samples that activate it the most. Figure~\ref{fig: L0A622} shows an example of the 64 inputs that activate L0A622 the most.

It is difficult to directly observe patterns from such images, and it is very prone to interpretability illusions. 
Therefore, we designed the interface as shown in Figure~\ref{fig: L0A622_interface}.

For a given dictionary feature, we examine the top-$k$ inputs that activate it the most among $n$ tokens and compute the following statistics over the $k$ input sequences/board states:

\begin{itemize}
    \item \textbf{Current player:}
        The pie chart in the first row first column shows the proportion of piece colors among the k inputs. 
    \item \textbf{Current move position:}
        The heatmap in the first row second column shows the number of times each tile is played in the k moves. 
    \item \textbf{Legal move positions:}
        The heatmap in the first row third column shows the number of times each tile is a legal move in the k board states. 
    \item \textbf{Board state:}
        The heatmap in the second row first column: for each tile in the k board states, take 1 if the cell has the same color as the current player ("own piece"), -1 if different color ("opponent piece"), and 0 if empty. 
        Summing over the k boards gives the total board state. 
    \item \textbf{Flip counts:}
        The heatmap in the second row second column shows how many times a piece is flipped on each cell among the k moves. 
    \item \textbf{Number of empty cells:} 
        The heatmap in the second row third column shows the number of empty cells in the k board states. 
    \item \textbf{Game length:}
        The bar plot in the third row shows the distribution of game lengths. 
\end{itemize}

We take $k$ as 2048 and $n$ as 1.2 million in our experiments. 
Each statistical plot reflects the behavior of one feature. 
The top-middle heatmap shows that among the 2048 inputs L0A622 is most interested in, all moves are played at position f-1. 
Therefore, we can interpret this as a "current move = f-1" feature.

\subsection{Othello-GPT and Language Models}

Our ultimate goal is to be able to reverse Transformer language models, using Othello models solely for conceptual verification. The Othello task is a decent simple substitute for language models, as its complexity makes it difficult to complete relying solely on memorization. The internal world of this "Othello language" is relatively rich, with many abstract concepts. The moderate complexity of this task gives it value as an initial attempt, compared to arithmetic tasks etc. which differ too much from language models, potentially failing to produce transferable conclusions. Meanwhile, it is not excessively complex so as to make training dictionaries or explanations overly difficult, with only 60 tokens in the vocabulary, and the lack of duplicate inputs to some extent enhances explainability, eliminating many easily conflated concepts.

A key difference between Othello and language models that we believe are worth noting in the context of dictionary learning is that the dictionary features are far more dense compared to language models: We find that Othello-GPT, especially in the middle layers, utilizes a great number of features to describe board states. As the game progresses and board states become more complex, the features are activated more frequently. This is consistent with existing observations of language models\citep{Elhage2022solu}.

\begin{figure}[h!]
    \centering
    \includegraphics[width=\linewidth]{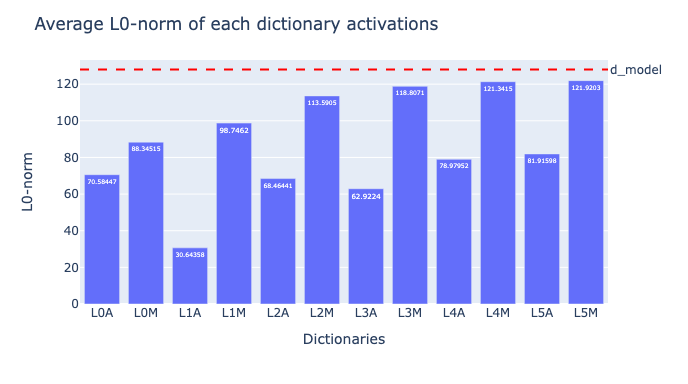}
    \caption{Average L0 norm of 10,000 tokens of each dictionary. Late MLPs in Othello-GPT exhibits higher L0-norm, close to d\_model = 128.}
    \label{fig: L0norm}
\end{figure}

As illustrated in Figure~\ref{fig: L0norm}, the L0-norm of decomposing the higher MLP layers sometimes approaches d\_model 128, indicating the dictionary decomposition is not very sparse. Compared to other work attempting dictionary decomposition of language models, such as \citet{Bricken2023monosemanticity}'s typically 10-20 sparse activations for a 512-d layer, our Othello dictionaries have much lower sparsity.

We largely attribute this to the fact that the Othello world is far simpler and has denser features compared to the linguistic world. We also believe our dictionary training implementation is likely not optimal.

\end{document}